\documentclass[letterpaper, 10pt, conference]{ieeeconf}      

\IEEEoverridecommandlockouts                              

\usepackage{times} 
\usepackage{amsmath} 
\usepackage{amssymb}  
\usepackage[dvipsnames]{xcolor}
\usepackage{microtype}
\usepackage[linesnumbered,ruled,lined]{algorithm2e}

\SetCommentSty{mycommentfont}

\usepackage{etoolbox}
\makeatletter
\patchcmd{\@algocf@start}
  {-1.5em}
  {-0.1em}
  {}{}
\makeatother

\usepackage[hidelinks]{hyperref}
\hypersetup{
    colorlinks=false,
    linkcolor=blue,
    urlcolor=blue,
    pdftitle={HiddenGems},
}

\usepackage{cleveref}
\usepackage{graphicx}

\def\HG/{\textsc{HiddenGems}}
\DeclareMathOperator{\abs}{abs}

\title{\LARGE \bf
\HG/: Efficient safety boundary detection with active learning
}

\author{
    Aleksandar Petrov,
    Carter Fang,
    Khang Minh Pham,
    You Hong Eng, \\
    James Guo Ming Fu,
    Scott Drew Pendleton
\thanks{All authors are affiliated with Motional, Inc., 100 Northern Avenue, 02210 Boston, MA. 
    Direct correspondence to \texttt{aleksandar@p-petrov.com}. } }

\begin{document}

\maketitle
\thispagestyle{empty}
\pagestyle{empty}

\begin{abstract}
    Evaluating safety performance in a resource-efficient way is crucial for the development of autonomous systems.
    Simulation of parameterized scenarios is a popular testing strategy but parameter sweeps can be prohibitively expensive.
    To address this, we propose \HG/: a sample-efficient method for discovering the boundary between compliant and non-compliant behavior via active learning.
    Given a parameterized scenario, one or more compliance metrics, and a simulation oracle, \HG/ maps the compliant and non-compliant domains of the scenario.
    The methodology enables critical test case identification, comparative analysis of different versions of the system under test, as well as verification of design objectives.
    We evaluate \HG/ on a scenario with a jaywalker crossing in front of an autonomous vehicle and obtain compliance boundary estimates for collision, lane keep, and acceleration metrics individually and in combination, with 6 times fewer simulations than a parameter sweep. 
    We also show how \HG/ can be used to detect and rectify a failure mode for an unprotected turn with 86\% fewer simulations. 
\end{abstract}

\section{Introduction}

Continuous safety assessment of the capabilities of safety-critical autonomous systems is a key part of their development process.
Some autonomous systems, such as autonomous vehicles (AVs), operate in complex uncertain environments and their behavior needs to generalize across novel circumstances.
This necessitates careful analysis of their performance.

No system can perform in all possible situations, so deployment is usually limited to an Operational Design Domain (ODD): a specification of the conditions under which the system is required to operate safely.
Furthermore, the system performance can be specified by one or more desired behaviors (\emph{rules}) each associated with a \emph{compliance metric} that measures to what extent the system exhibits the desired behavior.
Compliance metrics can be either binary (safe/unsafe) or continuous (degree of compliance).

The system developer then aims to produce a system that has safe behavior throughout the ODD for the given set of rules.
This often is a resource and time-consuming iterative process.
To assess the current state of the system under test and to inform the future development one needs to know in which parts of the ODD the system is currently compliant and where it needs further improvements.

The boundary between compliant and non-compliant behavior is therefore invaluable for the development process.
Such boundaries are usually in sensitive regions of the ODD where small perturbations tend to change the system's discrete state variables, resulting in large behavior differences.
Therefore, compliance boundaries provide valuable feedback by:
\begin{itemize}
    \item highlighting critical test cases for further root-cause analysis and fault identification;
    \item demonstrating compliance by showing that the compliance domain fully contains the ODD;
    \item tracking the performance change across the development process by means of compliance boundary shifts.
\end{itemize}

Evaluating regions of unsafe behavior and their boundaries cannot be safely performed on the physical system.
Therefore, system developers rely on an array of proxy methods. 
For autonomous robots, that is usually scenario-based simulation testing and validation \cite{scenarios}.
In particular, a developer could have a library of \emph{concrete scenarios} that correspond to points in the ODD against which they can evaluate the current state of the system.
Yet, testing single ODD points can lead to missing potential safety issues in their neighborhood.

Alternatively, one could partition the ODD into sets called \emph{logical scenarios} each containing similar concrete scenarios.
A logical scenario can be considered as a parameterized scenario with its concrete scenarios being indexed by a small number of parameters.
This results in an approximate embedding of the ODD onto a finite-dimensional parameter space.
Simulating the system on a grid of parameter values estimates its behavior across the logical scenario.
Such parameter sweeps can provide useful information but are often prohibitively expensive: a scenario with 4 parameters, 33 levels for each, taking 1 second per simulation would require almost two weeks of computation.
Most of the points on that grid are also typically non-informative as they lie far from the compliance boundary.
They are either too easy to handle by the system or completely ``unsavable''.

The pace of autonomous system development depends on the lag time from feature introduction to the identification of bugs and limitations.
Reducing the number of simulations needed can thus speed up the development process and reduce the associated costs and the environmental impact.
Consequently, there is a need for sample-efficient methods for discovering compliance boundaries in logical scenarios.

To address the challenge of finding compliance boundaries of a logical scenario in a sample-efficient manner, we propose the \HG/ framework.
Given a logical scenario with its set of parameters, a set of rules and corresponding compliance metrics, and an oracle, \HG/ sequentially selects the most informative concrete scenarios to query the oracle with.
These parameter values constitute the ``hidden gems'' of the parameter space: the ones that cost just as much to compute but are far more informative than the other sampling locations.
This constitutes an active learning problem \cite{settles2009active}.
The scenario queried is the one maximally reducing the \emph{ambiguity} (a notion of uncertainty) of the boundary for a single metric, for the combined compliance of two or more metrics, or for the identification of the most critical metric violation.
\HG/ then provides the user with an estimate of the compliance boundary and a set of concrete scenarios lying on both the compliant and non-compliant sides of the boundary.
The boundary estimate can then be used to evaluate which parts of the ODD meet the design objectives or for comparative analysis with other versions of the system.
Moreover, further investigation of the provided critical cases can guide the developer to identify the root-cause of the non-compliance.

The main contributions of this paper are:
\begin{itemize}
    \item a general sample-efficient method for critical test case identification and  compliance boundary estimation of logical scenarios with binary and continuous metrics;
    \item concrete implementations of \HG/ using both existing and novel active learning algorithms;
    \item extensions of \HG/ to combinations of violation metrics as well as importance hierarchies of them.
\end{itemize}

\section{Prior work}
\label{sec:prior_work}
Previous work finds examples of non-compliant behavior, without explicitly estimating the non-compliant region or its boundary.
Deep reinforcement learning techniques based on adaptive stress testing have been used to search for actor policies with a high likelihood for an event of interest \cite{koren2019efficient}. 
Importance sampling can obtain concrete scenarios resulting in accidents by utilizing statistical data on the likelihood of specific traffic situations \cite{okelly2018scalable}. 
However, AVs typically perform well under typical situations but it is edge cases that are truly challenging. 
To address this, Kl\"uck et al.\ and Beglerovic et al.\ propose genetic algorithms to explore the parameter space for concrete scenarios with low safety metric values \cite{kluck2019genetic,Beglerovic2017}.
Once such a scenario has been found, a local fuzzer can explore its neighborhood for other scenarios with non-compliant behavior \cite{li2020avfuzzer}.
Bayesian optimization techniques have also been used \cite{abeysirigoonawardena2019generating}.
These methods find instances of non-compliant behavior but do not focus on the compliance boundary.
In contrast, this paper, similarly to \cite{tuncali2018simulation,Mullins2017}, hones in on the scenarios located near the boundary.
Whereas \cite{tuncali2018simulation} employs covering arrays and simulated annealing and \cite{Mullins2017} requires a non-trivial meta-model, we use Gaussian processes and SVMs which come with theoretical guarantees on their performance, something particularly relevant for safety analysis.

The above methods target a single compliance measure, most frequently a collision metric.
Still, real-world systems need to comply with a number of legal requirements, good practices, and mission and comfort requirements which often cannot be satisfied simultaneously.
For instance, even if an AV has no admissible course of action, it still has to select which inadmissible action to take.
Handling both probabilistic and vague requirements during the design stage can be done via probabilistic model checking and fuzzy logic \cite{morse2017fuzzy}.
The Rulebooks framework provides a more comprehensive hierarchical setting permitting partial rule violations \cite{censi2019liability,collin2020safety}.
Rule violations are managed via a hierarchy of rules and violation metrics.
Violations of lower-priority rules (e.g. drive smoothly for AVs, maximize production output for industrial robots) are then preferred to violations of high-priority rules (e.g. prevent impact with other road users or objects, meet minimum manufacturing quality requirements).

\section{Problem statement}
\label{sec:problem_statement}

Given a logical scenario with $d$ parameters, its \emph{domain} $D\subset\mathbb{R}^d$ is the parameters indexing its concrete scenarios.
We further require a finite subset of the domain $C\subset D$, called \emph{query candidates}.
$C$ would often be a discrete grid of points in $D$ but other settings are also possible.

The behavior specification of the autonomous system is represented as a finite set of rules $\mathcal{R}$.
Each rule has an associated violation metric.
A binary violation metric is an unknown function $m: D\to\mathbb{B}$ inducing a partition of $D$ into a compliant subset $D^C_m=\{p\in D \mid m(p)=\mathtt{t}\}$ and a non-compliant subset $D^N_m=\{p\in D \mid m(p)=\mathtt{f}\}$.
Examples of binary violation metrics for AVs are non-collision with other road actors, stopping at a stop sign, and reaching the mission goal.
Similarly, a continuous violation metric is an unknown function $m': D\to\mathbb{R}$ and a threshold $t_{m'}$ inducing a compliant subset $D^C_{m'}=\{p\in D \mid m'(p)\leq t_{m'}\}$ and a non-compliant subset $D^N_{m'}=\{p\in D \mid m'(p)>t_{m'}\}$.
Some continuous metrics are excess velocity (thresholded against the speed limit) and excess acceleration (against a maximum comfort value). 
A continuous metric can be transformed into a binary one, though this prevents the learner from steering away from regions with violation values far from $t_{m'}$.
Setting $m(p)=\mathtt{f}$ and $m'(p)>t_{m'}$ to be the non-compliant values is arbitrary.
We denote the violation metrics for the scenario as $\mathcal{M}=\mathcal{M}_\mathbb{B}\cup\mathcal{M}_\mathbb{R}$, with $\mathcal{M}_\mathbb{B}\subset\mathbb{B}^D$ and $\mathcal{M}_\mathbb{R}\subset\mathbb{R}^D$ respectively being the sets of binary and continuous metrics.
We will assume there exists a bijection between $\mathcal{R}$ and $\mathcal{M}$ mapping each rule to its corresponding metric.

The objective of \HG/ is to select points in $C$ that are close to the boundary between $D^C_m$ and $D^N_m$ for a given metric $m$ by evaluating only a subset of them.
Evaluating a concrete scenario $p\in D$ amounts to querying an oracle for the metric values $m(p)$ for all $m\in\mathcal{M}$.
Driving simulators interfacing with the system under study can act as the oracle for AVs, while industrial robots simulators could be used for problems in manufacturing.

Depending on the objective and whether $\mathcal{R}$ is endowed with additional structure, we consider several modes of operation.
\begin{itemize}
    \item \textbf{Single rule:} When $\left|\mathcal{R}\right|=1$ there is only one metric $m$, either binary or continuous, and we are interested in the boundary between $D^C_m$ and $D^N_m$.
    \item \textbf{Collection of rules:} If $\left|\mathcal{R}\right|>1$ and we are interested in the compliance of the system relative to each separate rule. Hence we want to estimate the boundaries between $D^C_m$ and $D^N_m$ for all $m\in\mathcal{M}$.
    \item \textbf{Combination of rules:} If $\left|\mathcal{R}\right|>1$ and we are interested in \emph{total compliance}, i.e. the boundary between $D^C_{m_\text{tot}}$ and $D^N_{m_\text{tot}}$, where 
        \begin{equation}
            m_\text{tot}(p)=\bigwedge_{m\in\mathcal{M}} \bigtriangleup(m(p)),
            \label{eq:and_metric}
        \end{equation}
        with $\bigtriangleup$ being the compliance verification function
        \begin{equation*}
            \bigtriangleup(m(p)) = 
                \begin{cases}
                    m(p) & \text{if } m\in\mathcal{M}_\mathbb{B},\\
                    m(p)\leq t_m & \text{if } m\in\mathcal{M}_\mathbb{R}.
                \end{cases}
        \end{equation*}
        We obtain values for all metrics in $\mathcal{M}$ when evaluating a concrete scenario and can estimate all $|\mathcal{M}|$ individual boundaries, be it with lower precision.
        A high-quality estimate is obtained only for the total boundary.
    \item \textbf{Hierarchy of rules:} Often some rules are more important than others, as discussed in \Cref{sec:prior_work}.
        Hence, given a total order $<$ over the rules $\mathcal{R}$, we are interested in estimating the highest priority rule violated for every scenario in $D$.
        This can be represented as the boundaries where $m_<$ switches values:
        \begin{equation}
            \begin{aligned}
                m_< \colon D &\to \mathcal{M}\cup\{\top\} \\
                p &\mapsto
                    \begin{cases}
                        m & \text{if } \exists m, \text{s.t. } \neg\bigtriangleup(m(p)) \text{ and}\\
                          & \nexists m'>m, \text{s.t. } \neg\bigtriangleup(m'(p)),\\
                        \top & \text{otherwise},
                    \end{cases}
            \end{aligned}
            \label{eq:hierarchy_metric}
        \end{equation}
        with $\top$ the case when all rules are satisfied.
        We can also consider a partial order $\leq$ over $\mathcal{R}$ if for each equivalence class of rules induced by $=$ we take their combination (\Cref{eq:and_metric}) as a single level in the hierarchy.
\end{itemize}

\section{The \HG/ framework}

At iteration $i$, \HG/ picks $p\in C\backslash X_i$ maximizing an ambiguity function for the compliance metric.
$X_i$ are the previously selected points.
The definition of ambiguity is determined by the choice of active learning algorithm.
\Cref{sec:description_single_continuous,sec:description_single_binary} describe the ambiguity functions for existing and novel active learning algorithms for continuous and binary metrics.
While these work well in practice, \HG/ can use any other active learning algorithm.

Once $p$ is selected, the corresponding concrete scenario is processed by an oracle (e.g.\ a simulation tool) which provides the value $m(p)$ of every compliance metric $m\in\mathcal{M}$. 
\HG/ then updates its estimates of the response surface for each metric and picks a new point to simulate.
We assume a negligible cost of metric evaluation and response surface estimation compared to the cost of simulation.

When considering multiple rules, \HG/ takes turns picking the highest ambiguity point for each of them.
The exploration is further guided by controlling which query candidate points we consider.
This depends on whether the overall objective is to evaluate the collection, combination, or hierarchy of the rules, discussed in 
\Cref{sec:description_collection,sec:description_combination,sec:description_hierarchy}.

The \HG/ framework is general, i.e.\ it is agnostic of the learning algorithm choice.
Hence, one can modify the framework to fit their specific needs via the boundary and exploration selection functions $\xi_i^b, \xi_i^e$ in \Cref{algo:generic_exploration,algo:generic_exploration_multi}.

\subsection{Single rule with a continuous metric}
\label{sec:description_single_continuous}

We first address the compliance boundary of a single rule with a continuous associated metric $m_c$.
This is equivalent to finding the level set of $m_c$ at the threshold $t_{m_c}$.
Gaussian Processes (GPs) feature flexible response surface estimation and principled uncertainty quantification.
For this reason, GPs are used for active learning of level sets \cite{gotovos2013active,zanette2018robust,bryan2005active}.

A Gaussian Process is a collection of random variables, any finite subset of which is distributed as a multivariate Gaussian \cite{williams2006gaussian}. 
At every point $x$, the GP defines a mean $\mu(x)$ and variance $\sigma^2(x)$ of the values of the sampled functions at $x$.
The smoothness of the sampled functions (and their mean) is governed by the kernel choice $k(x,x')$.
Popular choices are the radial-basis function (RBF) and the Mat\'ern kernel.
Assuming a zero-mean prior, $N$ sampled locations $X=[x_1,\ldots,x_N]^\top$, measurements $Y=[y_1,\ldots,y_N]^\top$, and a kernel $k$, the posterior mean and variance are:
\begin{equation}
    \begin{aligned}
        \mu(x) &= \bar{k}(x)^\top \left(K + \sigma^2I\right)^{-1} Y \\
        \sigma^2(x) &= k(x, x) - \bar{k}(x)^\top \left(K + \sigma^2I\right)^{-1} \bar{k}(x), 
    \end{aligned}
    \label{eq:gp_fitting}
\end{equation}
with $\bar{k}(x) = \left[ k(x_1,x),\ldots,k(x_N,x)\right]^\top$ and $K_{i,j} = k(x_i,x_j)$.
The complexity of fitting a GP is $\mathcal{O}(N^3)$, but in practice it is negligible compared to the simulation time.

One popular approach to level set estimation is the eponymous LSE algorithm \cite{gotovos2013active}.
Given an accuracy parameter $\epsilon\in\mathbb{R}_{>0}$ and a confidence parameter $\delta\in(0,1)$, LSE
maintains confidence regions for all points in $C$ and updates them as more points are sampled. 
Points with confidence regions lying entirely above $t_{m_c}-\epsilon$ are labeled as higher than the threshold and those with confidence regions lying entirely below $t_{m_c}+\epsilon$ are labeled as lower.
The rest are considered yet unknown.
At iteration $i$, the next sample location is the unknown location with the highest ambiguity  $a_i^\text{LSE}(p)=\min\{\max (J_i(p))-t_{m_c}, t_{m_c}-\min (J_i(p))\}$, where
\begin{equation}
    J_i(p) = \bigcap_{i'=1,\ldots,i} \left[ \mu_{i'}(p) \pm \sqrt{\beta( i' )}\sigma_{i'}(p) \right],
    \label{eq:confidence_region}
\end{equation}
is the intersection of the confidence intervals at $p$ for all previous iterations, including $i$.
The posterior $\mu_{i'}$ and $\sigma_{i'}$ are obtained by conditioning on the first $i'$ samples and the interval half-size is 
$    \beta(i) = 2\log(|C|\pi^2i^2/(6\delta))$.
The algorithm terminates when there are no unknown points in $C$ left.
It is formally described in \Cref{algo:generic_exploration} with $\xi_i^b=\xi_i^e=a_i^\text{LSE}$, $T=\text{GP}$.
Note that LSE's objective is to minimize the error in the space of the range of $m_{c}$ rather than the error in locating the boundary (the domain space of $m_c$).
%

Hence, we propose a new algorithm, \emph{Gaussian Processes Regression with Boundary Exploitation and Space-Filling} (GPR-BE-SF), with an explicit exploration-exploitation trade-off. 
The exploration mode selects the most space-filling point, the maximizer of $a_i^\text{SF}(p) = \min_{x\in X_i} \left\|p-x\right\|$.
The exploitation mode of GPR-BE-SF refines the current boundary estimate by sampling a point close to it, that is, the minimizer of $a_i^\text{BE}(p) = \abs(\mu_i(p)-t_{m_c})$.
The chosen mode depends on how far we are in the exploration process: early on we favor exploration in order to find all non-compliant regions and to obtain a rough boundary estimate, while later we focus on refining this estimate.
To control the trade-off between the two modes, at iteration $i$ we sample on the border with probability $\tau(i)=\tanh(2i/N)$, where $N$ is the total number of iterations allocated.
Other functions with horizontal asymptotes 0 and 1 respectively at 0 and $N$ can be used in place of $\tanh$.
A small set of points $I\subset C$, usually a sparse grid, are simulated to initialize the first posterior.
The full procedure is presented in \Cref{algo:generic_exploration}, with $\xi_i^b=-a_i^\text{BE}$, $\xi_i^e=a_i^\text{SF}$, $T=\text{GP}$.

Note that $a^\text{SF}$ does not utilize the uncertainty information offered by the GP posterior.
Hence, it might unnecessarily sample in vast regions which are far from $t_{m_c}$ with high probability.
Therefore, we propose a modification of the algorithm, GPR-BE-LSE, inspired by LSE.
Keeping everything else the same, we only change the exploration objective from $a_i^\text{SF}$ to $a_i^\text{LSE}$ (\Cref{eq:confidence_region}).
GPR-BE-LSE corresponds to \Cref{algo:generic_exploration} with $\xi_i^b=-a_i^\text{BE}$, $\xi_i^e=a_i^\text{LSE}$, and $T=\text{GP}$.

\IncMargin{0.3em}
\begin{algorithm}[t]
    \small

    \SetKwInOut{Input}{input}\SetKwInOut{Output}{output}

    \Input{Candidates $C$, initial points $I\subset C$, method $T$, iterations $N$, boundary and exploration selection functions $\xi_i^b, \xi_i^e$, metric evaluation interface $m$ }
    \Output{Sample locations $X_N$ and values $Y_N$}

    \lFor{$i \leftarrow 1$ \KwTo $|I|$}{$x_i \leftarrow I_i, ~ y_i \leftarrow m(x_i)$ }
    \For{$i \leftarrow |I|+1$ \KwTo $N$}{
        $X_i \leftarrow \{x_1,\ldots,x_{i-1}\}$, $Y_i \leftarrow \{y_1,\ldots,y_{i-1}\}$\;
        \Switch{$T$\label{algline:fitstart}}{
        \uCase{GP}{
            Get $\mu_i, \sigma_i^2$ by fitting a GP with $X_i, Y_i$\tcp*{Eq. \ref{eq:gp_fitting}}
            }
        \Case{SVM}{
                Get $w_i$ by fitting an SVM with $X_i, Y_i$\tcp*{Eq. \ref{eq:SVM_definition}}
            }
        }\label{algline:fitend}
        $\tau \leftarrow \tanh(2i/N); ~ \text{sample } \bar{\tau} \sim U[0,1] $\;
        \eIf{$\bar{\tau}<\tau$}{
            $x_i \leftarrow \arg\max_{p\in C\backslash X_i} \xi_i^b(p), ~y_i \leftarrow m(x_i)$
        }{
            $x_i \leftarrow \arg\max_{p\in C\backslash X_i} \xi_i^e(p), ~y_i \leftarrow m(x_i)$
        }

    }
    $X_N \leftarrow \{x_1,\ldots,x_N\}$, $Y_N \leftarrow \{y_1,\ldots,y_N\}$\;
\caption{Single metric boundary detection}\label{algo:generic_exploration}
\end{algorithm}
\DecMargin{0.3em}

\subsection{Single rule with a binary metric}
\label{sec:description_single_binary}

Next, we address the case of a single rule with a binary compliance metric $m_b$.
A number of active learning algorithms for binary response functions have been proposed \cite{settles2009active}.
We outline three algorithms: two based on Support Vector Machines (SVMs) and one based on GPs.

An SVM is solving for the best separating hyperplane between two classes of samples or their embeddings via a kernel $k$ satisfying Mercer's condition \cite{cortes1995support}.
The best separation is achieved by maximizing the distance from the boundary to the samples closest to it.
Soft-margin SVMs relax this condition by allowing misclassification of some samples as a trade-off between the quality of the fit and the smoothness of the resulting boundary \cite{cortes1995support}.
Formally, a soft-margin SVM is defined as 
\begin{equation}
\begin{gathered}
    \min_{w, \zeta} \frac{1}{2} w^\top w + C' \sum_{i=1}^N \zeta_i,\\
    \begin{aligned}
        \text{subject to } &y_i (w^\top \phi(x_i) ) \geq 1-\zeta_i, \\
                           &\zeta_i \geq 0, i=1,\ldots,N,
    \end{aligned}
\end{gathered}
\label{eq:SVM_definition}
\end{equation}
with $w$ being the normal to the hyperplane in the embedding space, $\zeta$ a vector of slack variables, $y_i\in\{-1,+1\}$ the label for the sample $x_i$, $C'$ a regularization parameter for the smoothness-mislabeling trade-off, and $\phi$ an implicitly defined embedding function that must exist due to Mercer's theorem.

The simplest SVM-based active learning would be to pick the closest previously unsampled point to the current boundary estimate $w_i$, that is, the minimizer of the decision function $a_i^\text{DF}(p)=w_i^\top \phi(p)$.
Intuitively, this results in refining the boundary estimate and is also supported by theory, as it approximates choosing the point that halves the volume of the space of hyperplanes that fit the data \cite{tong2001support,freund1997selective}.
We will refer to this procedure as SVM-DF.
In this case, there is no additional explicit exploration, hence SVM-DF corresponds to \Cref{algo:generic_exploration} with $\xi_i^b=\xi_i^e=-a_i^\text{DF}, T=\text{SVM}$.
Alternatively, we can combine sampling the decision function minimizer together with the space-filling explorer, which we would call SVM-DF-SF.
This corresponds to \Cref{algo:generic_exploration} with $\xi_i^b=-a_i^\text{DF}, \xi_i^e=a_i^\text{SF}, T=\text{SVM}$.

We can use Gaussian Processes Classification (GPC) in a similar way.
GPC is a probabilistic classification algorithm that uses a latent GP mapped to probabilities with, e.g.\ the logistic function $f(x)=(1+e^{-x})^{-1}$, resulting in probability $\pi(p)$ of a point $p$ being from the positive class.
Hence the border at iteration $i$ is the 0.5 level set of $\pi_i$, or equivalently, the 0 level set of $\mu_i$ \cite{williams2006gaussian}.
We will call this procedure GPC-P-SF.
Concretely, that is \Cref{algo:generic_exploration} with $\xi_i^b=-a_i^\text{P}, \xi_i^e=a_i^\text{SF}, T=\text{GP}$, where $a_i^\text{P}(p) = \abs(\mu_i(p)-0)$.

\subsection{Collection of rules}
\label{sec:description_collection}

\IncMargin{0.1em}
\begin{algorithm}[t]
    \small
    \SetKwInOut{Input}{input}\SetKwInOut{Output}{output}

    \Input{\textls[-1]{Candidates $C$, initial points $I\subset C$, algorithm $A$, iterations $N$, selection functions $\xi_i^{b,m}, \xi_i^{e,m}$ and methods $T^m$ for all metric evaluation interfaces $m\in\mathcal{M}$} }
    \Output{Sample locations $X_N$ and values $Y_N^m, \forall m\in\mathcal{M}$}
    
    \lFor{$i \leftarrow 1$ \KwTo $|I|$}{$x_i \leftarrow I_i, ~ y_i^m \leftarrow m(x_i), \forall m\in\mathcal{M}$}
    \For{$i \leftarrow |I|+1$ \KwTo $N$}{
        $X_i \leftarrow \{x_1,\ldots,x_{i-1}\}$, $Y_i^m \leftarrow \{y_1^m,\ldots,y_{i-1}^m\}, \forall m\in\mathcal{M}$\;
        
        \For{$m \in \mathcal{M}$}
            {Fit $X_i, Y_i^m$ with $T_m, \xi^{b,m}_i, \xi^{e,m}_i$\tcp*{Alg. \ref{algo:generic_exploration}, lines \ref{algline:fitstart} to \ref{algline:fitend}}}
            
        $\bar{m} \leftarrow \mathcal{M}[1+(i\mod|\mathcal{M}|)]$\tcp*{for a fixed sequence on $\mathcal{M}$}
        
        \Switch{$A$}{
        \lCase{M-TT}{$Q_i \leftarrow C_i/X_i$}
        \lCase{M-C\hspace{1.15ex} }{$Q_i \leftarrow (C_i/X_i) \cap \bigcap_{m\in\mathcal{M}} D^C_{m,i}$}
        \lCase{M-H\hspace{1.08ex} }{$Q_i \leftarrow (C_i/X_i) \cap \bigcap_{m'>\bar{m}} D^C_{m',i}$}
        }
        $\tau \leftarrow \tanh(2i/N); ~ \text{sample } \bar{\tau} \sim U[0,1] $\;
        \eIf{$\bar{\tau}<\tau$}{
            $x_i \leftarrow \arg\max_{p\in Q_i} \xi_i^{b,\bar{m}}(p)$
        }{
            $x_i \leftarrow \arg\max_{p\in C/X_i} \xi_i^{e,\bar{m}}(p)$
        }
        $y_i^m \leftarrow m(x_i), \forall m\in\mathcal{M}$

    }
    $X_N \leftarrow \{x_1,\ldots,x_N\}$, $Y_N^m \leftarrow \{y_1^m,\ldots,y_N^m\}, \forall m\in\mathcal{M}$\;
\caption{Multi-metric boundary detection}\label{algo:generic_exploration_multi}
\end{algorithm}
\DecMargin{0.1em}

So far we have handled the case for finding the compliance boundary for a single rule.
However, in practice, an autonomous system often needs to comply with a collection of rules $\mathcal{R}$.
A single-metric procedure as proposed in \Cref{sec:description_single_continuous,sec:description_single_binary} can be independently applied for each rule, though that would require $|\mathcal{R}|$ times the number of samples.
Instead, samples chosen for the exploration of one of the metrics can be used to update the estimates for the other metrics as well.
Hence, we employ a turn-taking algorithm called M-TT.
Given a set of metrics $\mathcal{M}$, M-TT iterates which learner it queries for the next location but evaluates the metrics for all rules and updates the estimates for all learners.
This holds even if the individual learners are different and is described in \Cref{algo:generic_exploration_multi} with $A=\text{M-TT}$.

\subsection{Combination of rules}
\label{sec:description_combination}
Often one prefers higher-accuracy estimates for the boundary of all rules being compliant instead of the compliance domain of individual rules.
This amounts to the boundary for the binary-valued metric $m_\text{tot}$ as in \Cref{eq:and_metric}.
Na\"{i}vely, $m_\text{tot}$ is like any other binary metric and the algorithms proposed in \Cref{sec:description_single_binary} can be used.
However, in this way we throw away information from the continuous metrics $\mathcal{M}_\mathbb{R}$ which could otherwise steer the sampling away from regions with metric values far from their thresholds.
Instead, we propose M-C: an algorithm using M-TT as a base but restricting exploitation to points in $\cap_{m\in\mathcal M}D^C_{m,i}$, the current estimate of the compliant domain for $m_\text{tot}$ (\Cref{algo:generic_exploration_multi} with $A=\text{M-C}$).
Note that we still pick space-filling points in the non-compliant regions, preventing an initial bad estimate from irreversibly biasing the exploration.
We can also still obtain boundary estimates for all metrics, be it of lower resolution.

\subsection{Hierarchy of rules}
\label{sec:description_hierarchy}
Finally, we consider being presented with an order $<$ over the rules and the question of which is the highest violated rule for a given scenario. 
In particular, we are interested in the boundaries where $m_<$ (\Cref{eq:hierarchy_metric}) changes values.
Again, a na\"{i}ve approach would be applying M-TT and then computing the value of $m_<$ for all sampled locations.
However, that results in unnecessary sampling around lower-priority boundaries known to be non-compliant to a higher-priority rule.
Hence, we propose M-H, an algorithm that restricts the boundary detection of lower-priority metrics only to the intersection of the current estimates for the compliant domains for the higher-priority rules.
This is \Cref{algo:generic_exploration_multi} with $A=\text{M-C}$.
Similar to M-C, we still pick space-filling points and also obtain estimates for all metrics as a byproduct.

\section{Experiments}
\label{sec:experiments}

\begin{figure}
    \includegraphics[width=\linewidth]{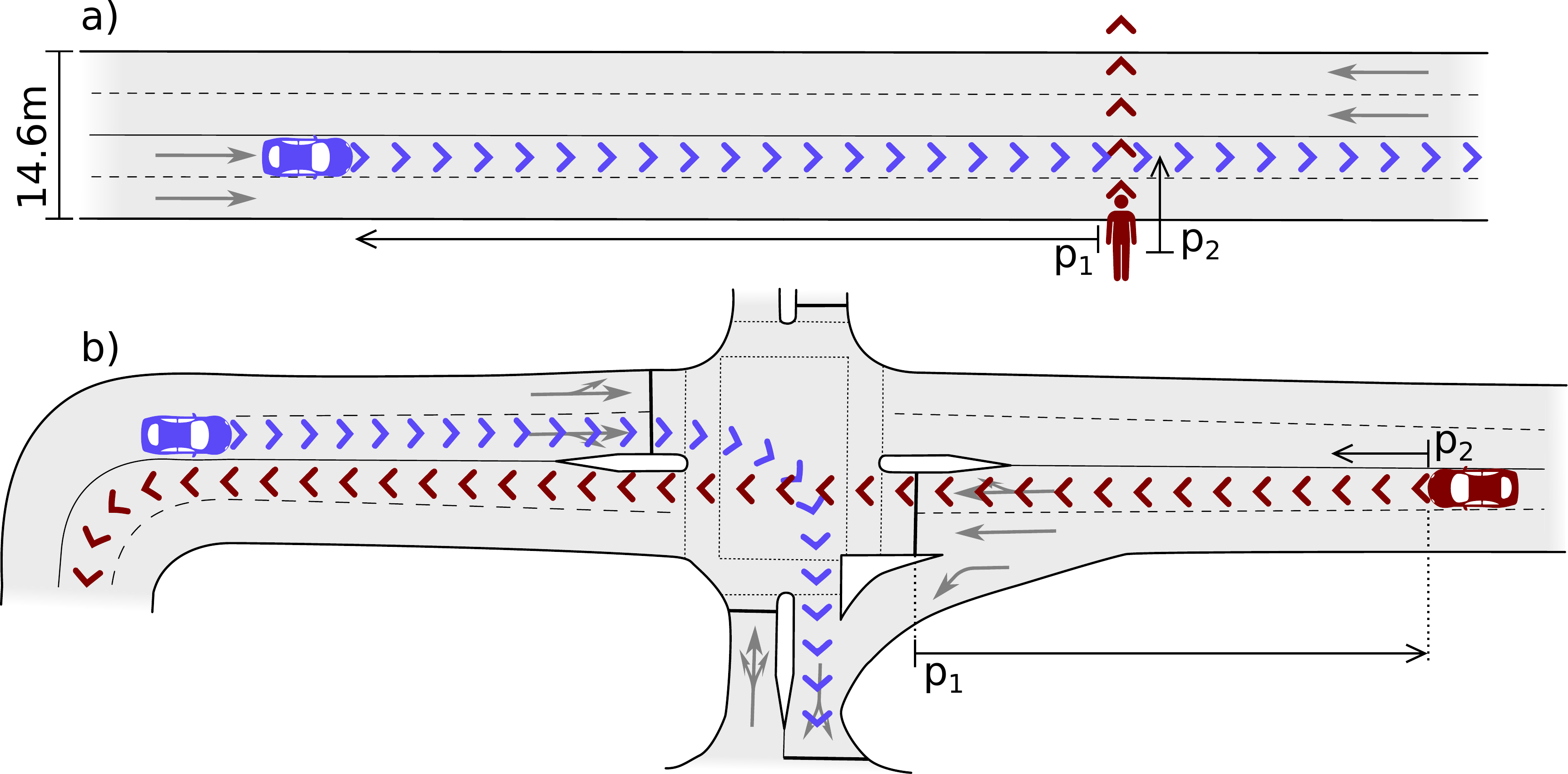}
    \caption{The jaywalker scenario (a) and the unprotected turn scenario (b).}
    \label{fig:scenarios}
\end{figure}

\begin{figure*}[t]
    \centering
    \includegraphics[width=\linewidth]{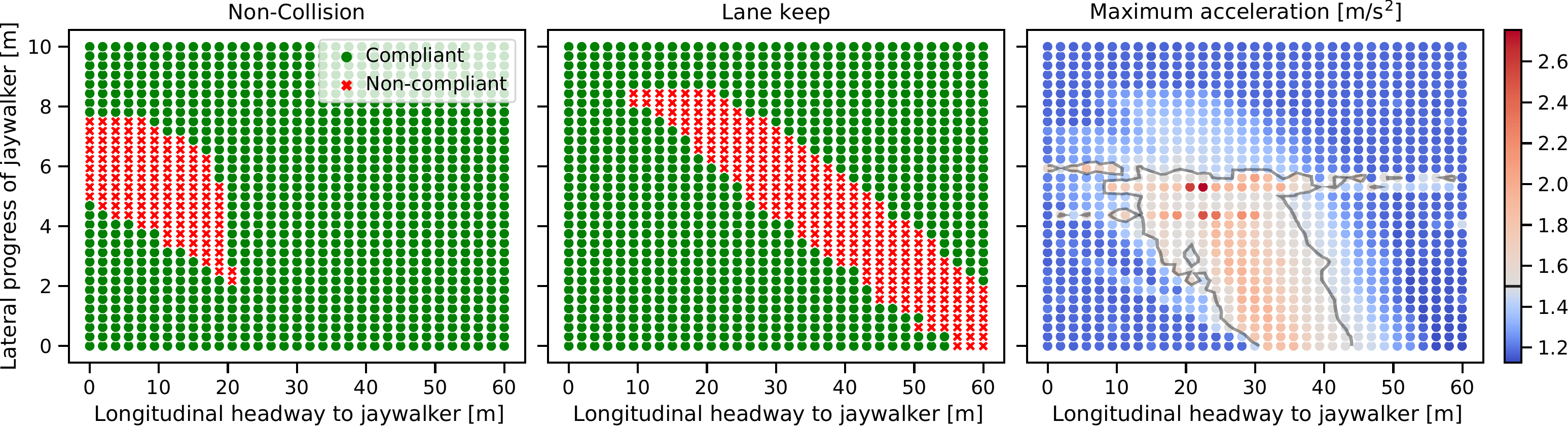}
    \caption{Ground truth values for the Non-Collision, Lane keep, and Maximum acceleration metrics in the jaywalker scenario.}
    \label{fig:ground_truth}
\end{figure*}

In this section, we provide examples of how the \HG/ framework can be applied in the development process of safety-critical systems, such as AVs.
The results are for illustration purposes only and were performed with a simplified planner and simulator, thus are not indicative of the performance of vehicles deployed on public roads.

We will first consider a jaywalker scenario.
The ego vehicle (whose performance we are studying) is driving on the left lane of a right-hand traffic carriageway with four traffic lanes, two in each direction.
A jaywalker appears in front of the ego and crosses the road.
We parameterize the longitudinal headway ($p_1\in[0\text{m},60\text{m}]$) and the lateral progress ($p_2\in[0\text{m},10\text{m}]$) of the jaywalker at the moment of first detection (\Cref{fig:scenarios}a).
Hence, the domain of the logical scenario is $D=[0\text{m},60\text{m}]\times [0\text{m},10\text{m}]$.
%
The jaywalker crosses the track of the ego at $p_2=6.24\text{m}$.
This scenario is representative of actual challenging conditions witnessed by AVs \cite{UBER}.

We are interested in three metrics, in order of importance:
\begin{enumerate}
    \item Primary safety rule metric $m_\text{col}\in\mathbb{B}^D$: avoid collision with the pedestrian. Violated if the minimum distance between the bounding boxes of the ego and the pedestrian is less than 15cm (buffer reflecting uncertainties).
    \item Traffic rule metric $m_\text{lane}\in\mathbb{B}^D$: do not leave the current lane. Violated if the bounding box of the ego is 25cm or more outside its own lane.
    \item Comfort rule metric $m_\text{acc}\in\mathbb{R}^D$: drive comfortably for the passenger. Violated if the absolute acceleration of the ego exceeds $t_{m_\text{acc}}=\text{1.5m/s}^2$ (passengers start experiencing discomfort around this value \cite{acceleration}).
\end{enumerate}
We also consider a binary version $m_\text{bacc}$ of $m_\text{acc}$ that we define as $m_\text{bacc}(p)=(m_\text{acc}(p)  \leq t_{m_\text{acc}})$.

The query candidates $C$ are a $33\times 33$ grid over $p_1$ and $p_2$, with a $6\times 6$ initialization grid $I$.
The coordinates of the points are normalized to the $[0,1]$ range.
As ground truth, we evaluate the three metrics on all 1089 candidate points.

The ground truth values for the three metrics are in \Cref{fig:ground_truth}.
When the jaywalker appears close to the ego and depending on their lateral location, the ego might not have enough time to react so a collision occurs.
For lateral starting positions far left or right of the ego, the jaywalker passes either before or after the ego, averting a collision.
The plots for $m_\text{lane}$ and $m_\text{acc}$ show that the ego successfully avoids collision by decelerating rapidly, changing lanes, or both.
If the jaywalker appears far away and in front of the ego or to the left, they have crossed the road by the time the ego crosses their path so no action beyond mildly decelerating is required.

To measure the quality of the boundary detection, we use Balanced Accuracy Score on Border: Balanced Accuracy Score computed only over the points lying on the compliance border for the ground truth.
A point $p$ is considered to be on the border for metric $m$ if for some $p'$ in its $3\times 3$ neighborhood it holds that $\bigtriangleup(m(p))\neq\bigtriangleup(m(p'))$.

\begin{figure}
    \includegraphics[width=\linewidth]{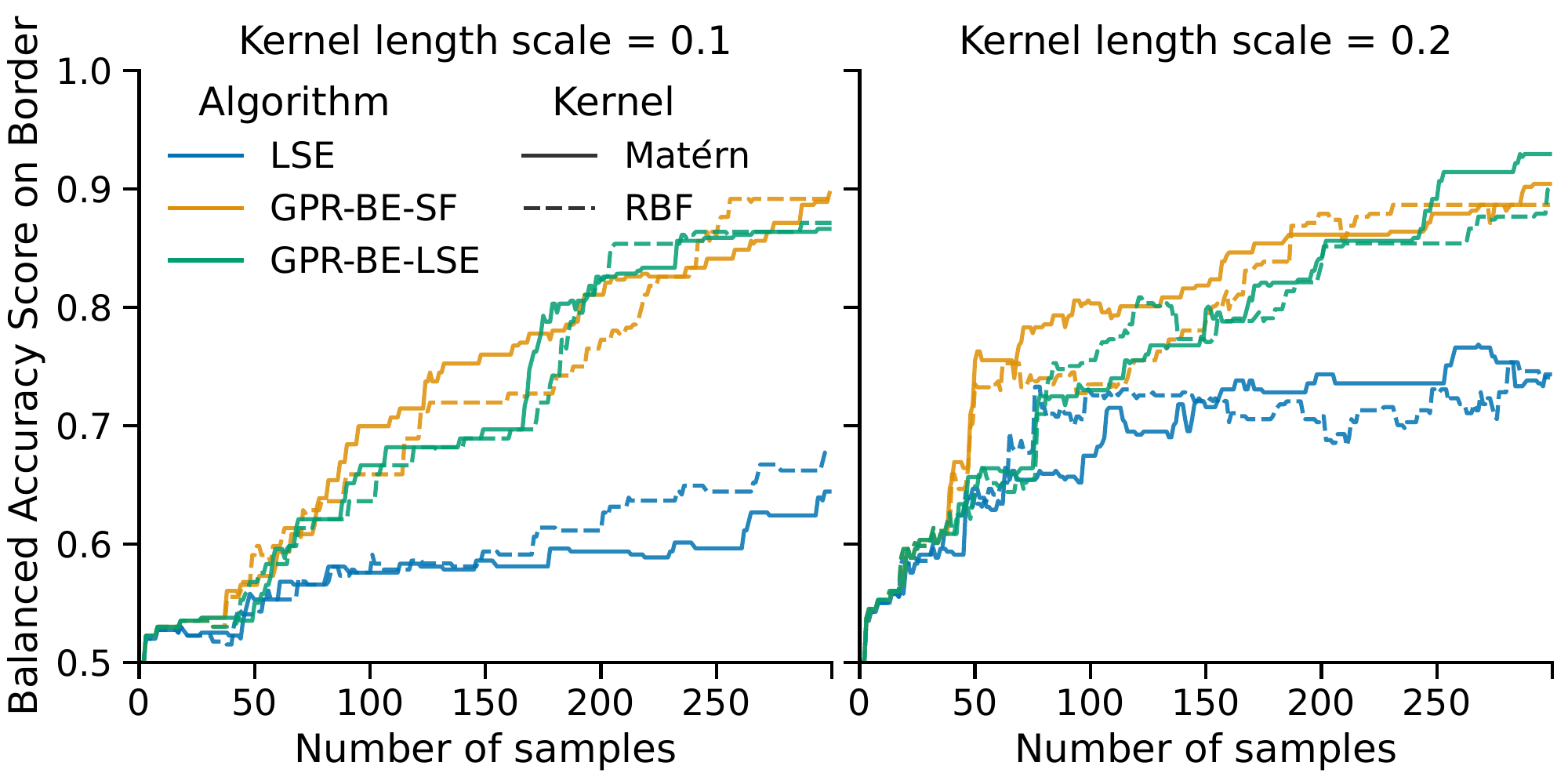}
        \caption{Comparison of the algorithms for a single continuous metric.}
    \label{fig:comp_single_real}
\end{figure}

\begin{figure}
    \includegraphics[width=\linewidth]{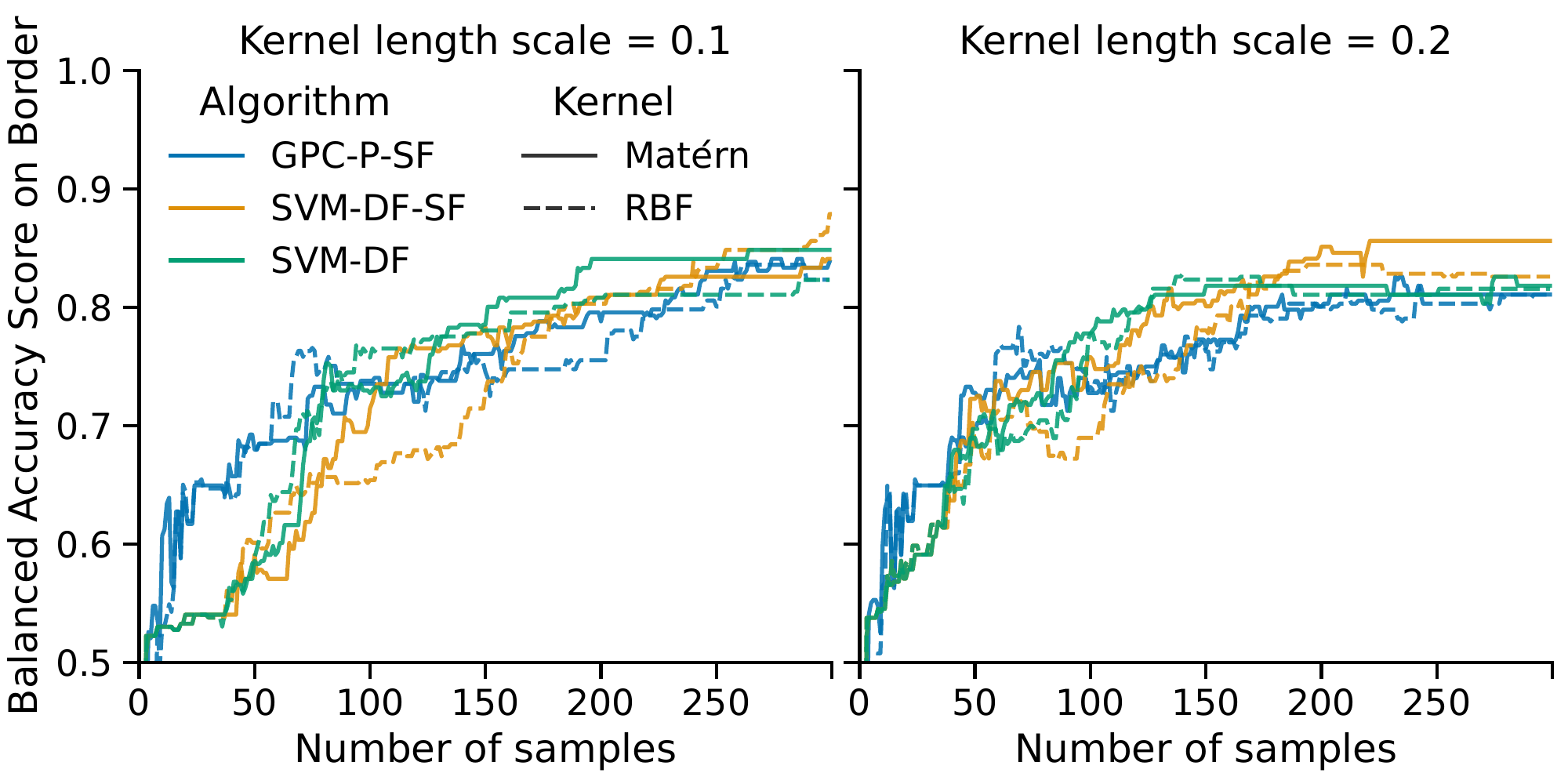}
    \caption{Comparison of the algorithms for a single binary metric.}
    \label{fig:comp_single_binary}
\end{figure}

\begin{figure*}[t]
    \centering
    \includegraphics[width=\linewidth]{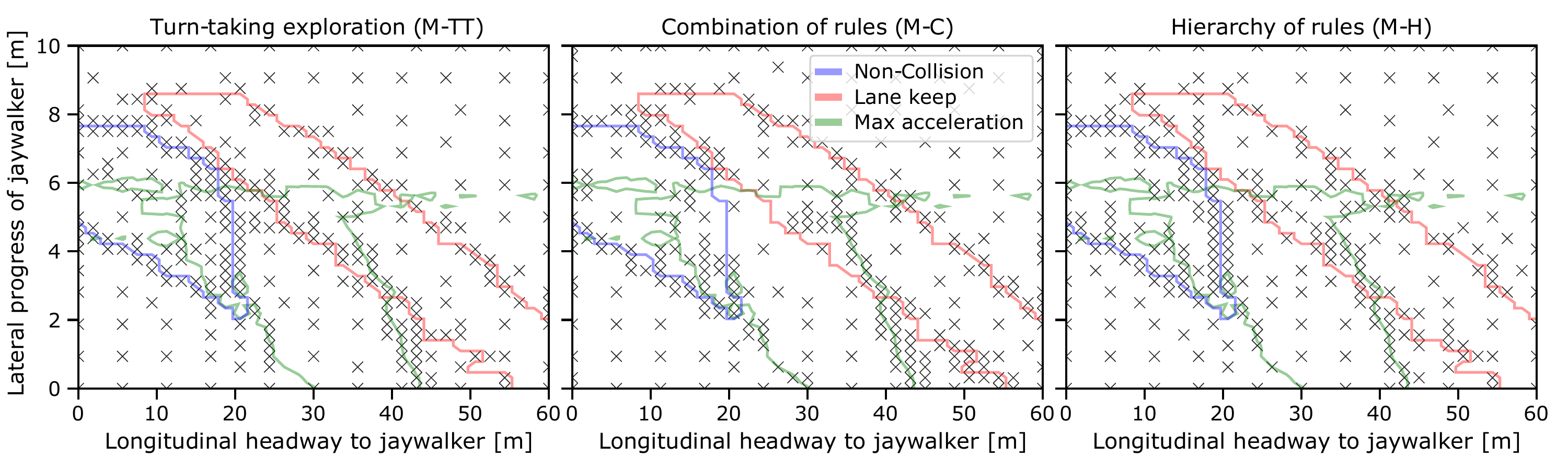}
    \caption{Selected sampling locations for the multiple metrics case with the ground truth compliance boundaries plotted for the three metrics.}
    \label{fig:samples_multimetric}
\end{figure*}
 
\subsection{Compliance boundary detection for individual rules}

\Cref{fig:comp_single_real} compares the three algorithms proposed in \Cref{sec:description_single_continuous} on the $m_\text{acc}$ metric for length scales 0.1 and 0.2 and with both the RBF kernel and the Mat\'ern kernel (smoothness parameter 2.5) for a maximum of 300 iterations.

\begin{figure}
    \centering
    \includegraphics[width=\linewidth]{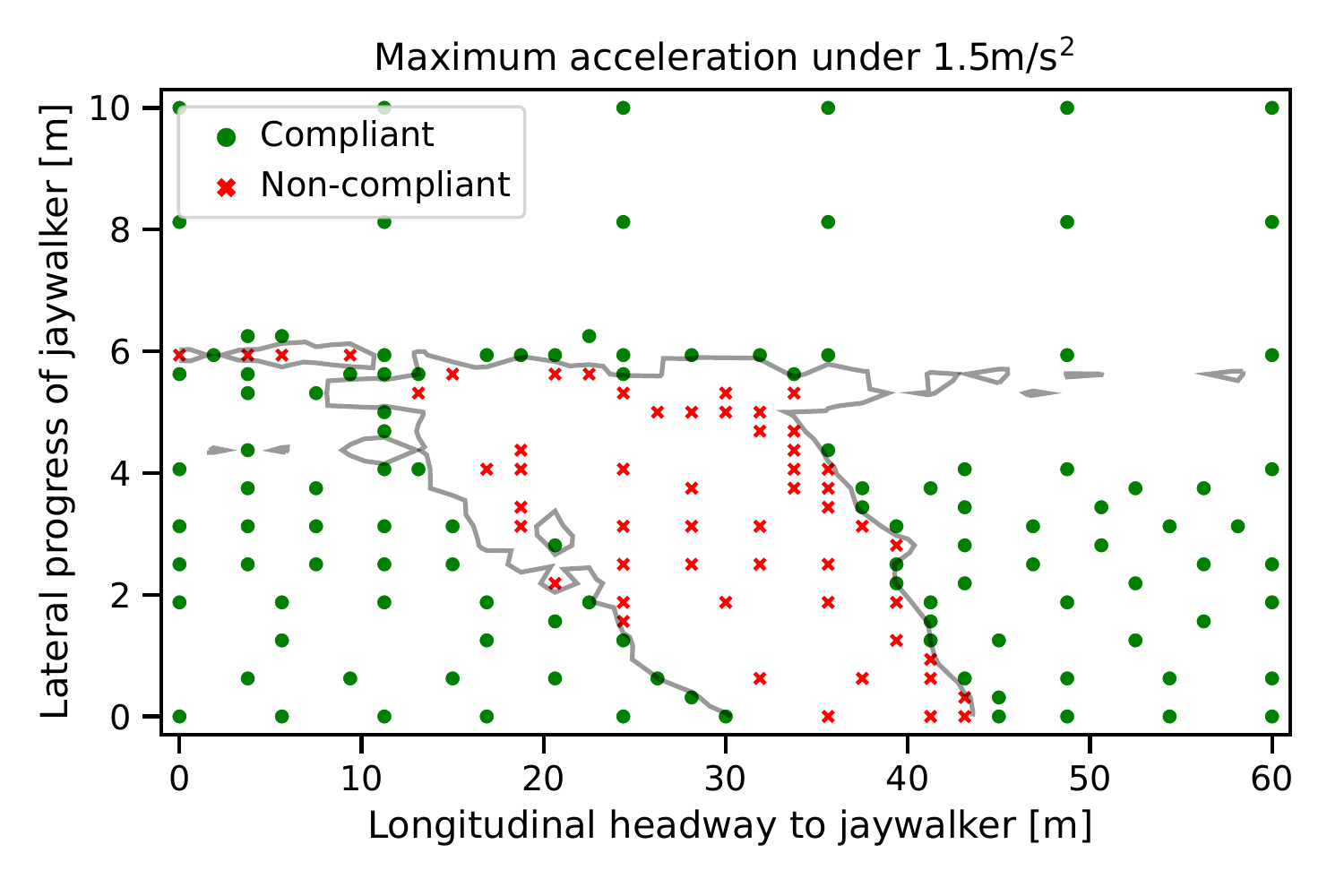}
        \caption{First 170 selected scenarios for    GPR-BE-LSE with Mat\'{e}rn, 0.2.}
    \label{fig:real_exploration_locations}
\end{figure}

For this task, GPR-BE-SF and GPR-BE-LSE both outperform LSE across the choices for length scale and kernel.
The Mat\'ern kernel appears to be performing slightly better than RBF, though this could be specific to the scenario.
Both GPR-BE-SF and GPR-BE-LSE achieve about 90\% correct classification on the boundary by just sampling a quarter of the points.
This is already a significant reduction in the time and resources needed for simulation.
In practice, however, terminating only after about 170 iterations provides sufficiently refined boundaries for development purposes, which is a reduction of the computational cost of about 6.4 times (see an example in \Cref{fig:real_exploration_locations}).

It appears that perfect quality is approached as the number of samples nears 1089, the total number of candidate points.
However, as the algorithm and its hyperparameters induce a family of models with specific smoothness properties, perfect boundary reconstruction will not be possible if the true response is not from this family.
Hence, hyperparameter selection should be performed using domain knowledge.

\Cref{fig:comp_single_binary} compares the three binary metric algorithms from \Cref{sec:description_single_binary}
on $m_\text{bacc}$.
For the SVM-based algorithms, $C'$ was set to 10.
Unlike the continuous case, all binary algorithms exhibit similar performance.
GPC-P-SF scores slightly lower than the SVM-based algorithms but the difference is negligible.
For the binary case, we too can achieve about 80\% balanced accuracy score on the border by sampling less than a quarter of the candidate points.
The performance of the binary case appears to be slightly lower than the one for the continuous case, signaling that using metrics with continuous values, when available, is beneficial.

\subsection{Compliance boundary detection for multiple rules}

\Cref{sec:description_collection,sec:description_combination,sec:description_hierarchy} introduced the M-TT, M-C, and M-H algorithms, respectively for the collection, combination and hierarchy settings.
\Cref{fig:samples_multimetric} shows how the three algorithms result in different sections of the compliance boundaries of $m_\text{col}$, $m_\text{lane}$, and $m_\text{acc}$ being explored, as desired.
The learner for $m_\text{col}$ is GPC-P-SF with the Mat\'{e}rn kernel, for $m_\text{lane}$ is the SVM-DF-SF with RBF, and for $m_\text{acc}$ is GPR-BE-LSE with Mat\'{e}rn, all with length scale 0.2 and ran for 300 iterations.

Comparing the sampling locations selected by M-C and M-TT, we observe that M-C samples more on the non-overlapping boundaries and less on parts of the boundaries in the non-compliant domains of other rules.
For example, the part of the boundary of $m_\text{col}$ that is in $D^N_{m_\text{acc}}$ is not as extensively explored as by M-TT.
The same holds for the segment of $m_\text{lane}$'s boundary that falls in $D^N_{m_\text{acc}}$.
Hence, M-C indeed prioritizes the border of $m_\text{tot}$ which is precisely the boundary of the union of the three non-compliant regions.

Similarly, comparison with M-TT shows that M-H sampled extensively the border of the highest-importance metric $m_\text{col}$, followed by the one for $m_\text{lane}$.
This is evident from the increased sampling in the overlapping region of the boundaries of $m_\text{col}$ and $m_\text{lane}$ with $D^N_{m_\text{acc}}$ and the reduced sampling in the region of the boundary of $m_\text{acc}$ that intersects the non-compliant domains of the other two metrics.

\subsection{Compliance boundary detection as a development tool}
\label{sec:example_cetran}

\begin{figure}
    \includegraphics[width=\linewidth]{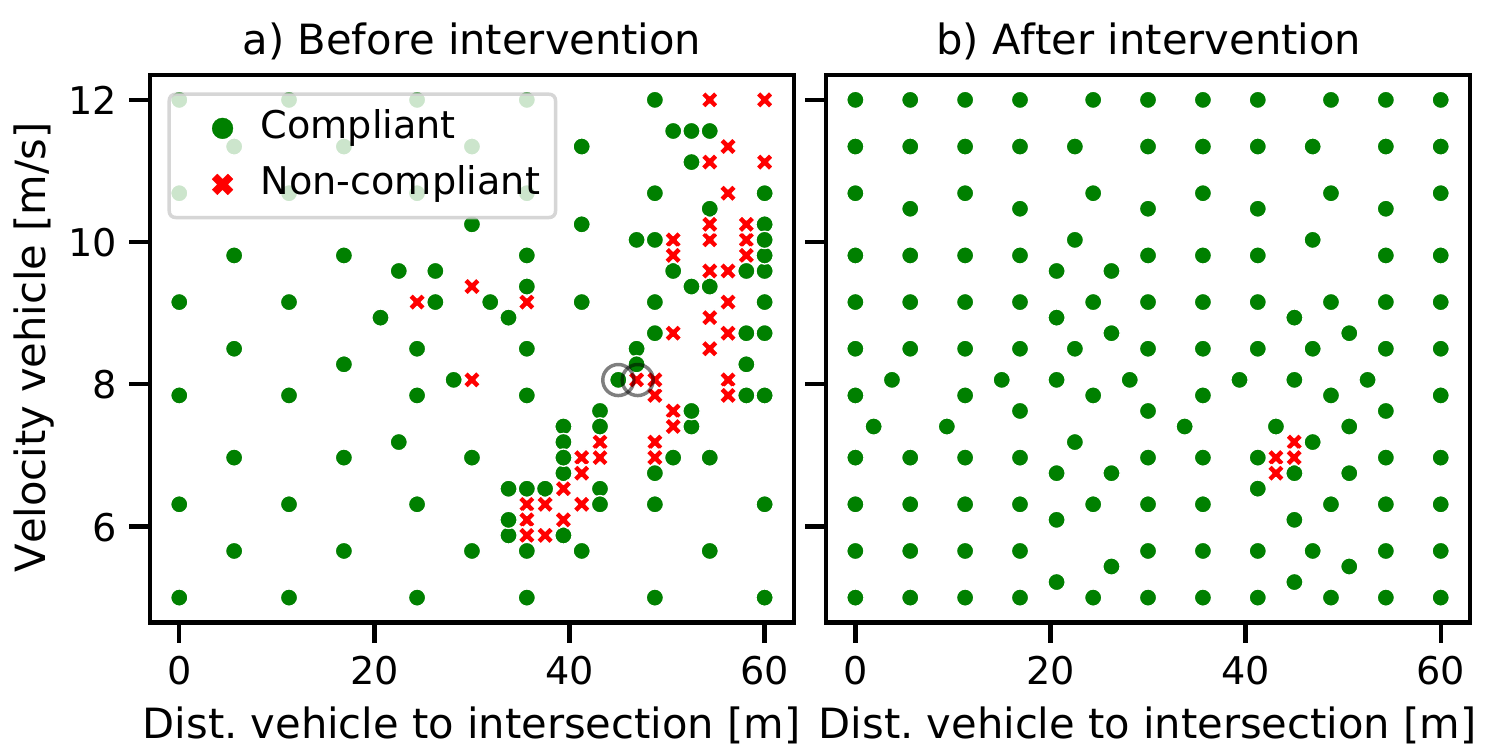}
    \caption{Points on the non-collision compliance boundary for the unprotected turn scenario before and after investigating and rectifying the unsafe phenomenon.}
    \label{fig:cetran_before_after}
\end{figure}

\begin{figure}
    \includegraphics[width=\linewidth]{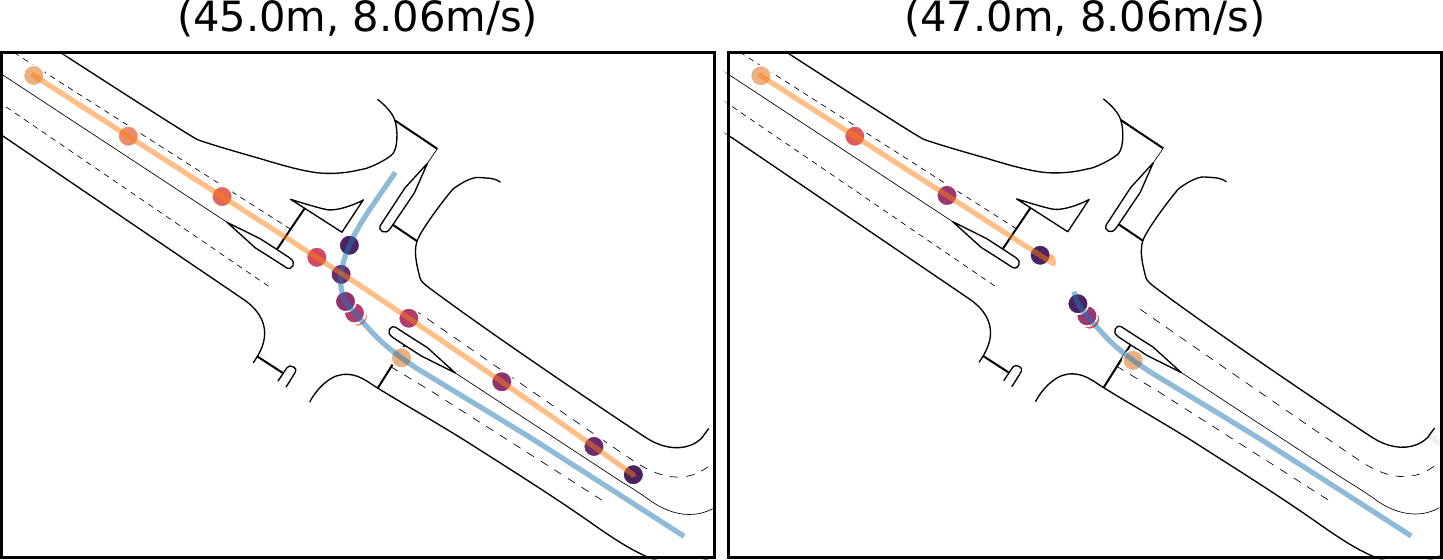}
    \caption{Plot of the tracks of the ego (blue) and agent (orange) for the two marked cases in \Cref{fig:cetran_before_after}. Markers with the same color designate locations at which the two vehicles were at the same time.}
    \label{fig:tracks}
\end{figure}

We present a further example from our practice.
We studied the behavior of the simplified planner for a turn in an unsignalled intersection, which is considered a difficult AV behavior to master.
The scenario in question is set in left-hand traffic where the ego attempts an unprotected right turn. 
As it enters the intersection, the ego detects an oncoming vehicle (agent) and must decide whether to yield or to attempt to clear the intersection before the agent.
We parameterize the distance of the agent to the intersection at the moment of first detection ($p_1\in[0\text{m},60\text{m}]$) as well as the agent's constant speed ($p_2\in[5\text{m/s},12\text{m/s}]$), see \Cref{fig:scenarios}b.
If the vehicle is far away or is moving slowly, the ego can  clear the intersection before it.
Similarly, if it appears very close or is very fast, it will clear the intersection before the ego intersects its path.
In all other cases, the ego would have to brake and yield.

We ran an exploration of the collision metric $m_\text{col}$ with GPC-P-SF (Mat\'{e}rn, 0.2, 150 iterations) and obtained the points on the boundary estimate plotted in \Cref{fig:cetran_before_after}a.
Surprisingly, the current state of the system handled vehicles appearing very close and moving with high velocities but failed for much simpler cases.
To understand this behavior, we performed a closer inspection of nearby scenario pairs lying on different sides of the border.
For example, the tracks of the marked pair of concrete scenarios are shown in \Cref{fig:tracks}.
In the compliant case, the ego stops at the appropriate location and yields to the agent, as required.
In the non-compliant case, with the agent starting just two meters further away, the behavior is drastically different: the ego slows down but does not stop, which results in it being hit by the agent.
A close inspection of the two marked scenarios determined the cause of this behavior to be an inappropriate choice of the parameters governing the intersection navigation behavior: the maximum time until the agent enters the intersection for which the ego yields was set too low.
Increasing this parameter from 4 to 10 seconds, and running \HG/ once again to verify the effect of the changes demonstrated a much improved compliant domain (\Cref{fig:cetran_before_after}b).

The area of the non-compliant region in \Cref{fig:cetran_before_after}b is very small relative to the total area.
If only a single or a handful of pre-determined concrete scenarios were employed, likely none of them would have been in the non-compliant region, incorrectly indicating safe behavior for this logical scenario.
On the other hand, a parameter sweep over the same $33\times 33$ grid, would require more than 7 times the number of scenario evaluations, almost all of which would be very easy to handle and far from the compliance boundary.
Hence, \HG/ brings the best from both worlds: successfully detecting the non-compliant region without performing unnecessary sampling.

\section{Discussion and conclusions}

The experiments in \Cref{sec:experiments} demonstrated that \HG/ discovers the compliance boundaries for rules with binary and continuous violation metrics, as well as combinations of them, with fewer simulations relative to a parameter sweep.
The insight provided by \HG/ can be indispensable for the development process.
For instance, pairs of scenarios exemplifying current issues present invaluable context for performing root cause analysis and rectification.
\HG/ can also be leveraged as a systems engineering tool to monitor the development progress towards satisfaction of complex requirements, namely, by tracking the evolution of $D^C_{m_\text{tot}}$ or $D^C_{m_<}$.
Through tracking the evolution of the whole domain $D$ rather than one or few single points in it, it provides an advantage to the traditional scenario library testing strategy.

While \HG/ was developed with the challenges of developing AVs in mind, the technique is applicable to any autonomous system that can be tested in simulation.
The use-case is not limited to safety evaluation in simulation and can be also applied to experimental setups (e.g. materials design, \emph{in vitro} studies) where any combination of parameters can be physically tested in a safe manner.
Even though the examples considered here were restricted to two dimensions, the same techniques can be employed in higher dimensions as well.
The single metric algorithms in \Cref{sec:description_single_continuous,sec:description_single_binary} can also be replaced with other active learning algorithms, as \HG/ is agnostic of the specific algorithm choice.

Other extensions of \HG/ are left as future work.
A hyperparameter-tuning procedure is needed for all algorithms presented in the current work.
Presenting the results to a developer for logical scenarios with more than three parameters is also an open question.
Similarly, employing \HG/ in a continuous integration pipeline requires automatic determination if a new feature constitutes improvement, regression, or neither.
Further research in incorporating batch sampling algorithms can provide speedup via parallel simulation execution.
Finally, statistical guarantees on the performance of the boundary detection algorithms can strengthen the confidence in \HG/'s capabilities.

\vfill
\bibliographystyle{IEEEtran}
\bibliography{bibliography}



\end{document}